\documentclass[letterpaper, 10 pt, conference]{ieeeconf}  

\IEEEoverridecommandlockouts                              

\usepackage{graphics} 
\usepackage{graphicx}
\usepackage{epsfig} 
\usepackage{epstopdf} 
\usepackage{amsmath} 
\usepackage{amssymb}  

\usepackage{microtype}
\usepackage[usenames,dvipsnames,table]{xcolor}
\usepackage{commath}
\usepackage{todonotes}
\usepackage{bm}
\usepackage{mathtools}

\usepackage[utf8]{inputenc}
\usepackage[english]{babel}
\PassOptionsToPackage{hyphens}{url}\usepackage[colorlinks, citecolor={blue}, linkcolor={blue}]{hyperref}
\usepackage{import}
\usepackage{paralist}
\usepackage{algpseudocode}
\usepackage{algorithm}
\usepackage{gensymb}
\usepackage{dsfont}
\usepackage{siunitx}
\usepackage[bottom]{footmisc}
\usepackage{framed}
\usepackage{balance} 
\usepackage{caption}
\usepackage[skins]{tcolorbox}  
\usepackage{subcaption}

\usepackage{todonotes}

\DeclareCaptionFont{mysize}{\fontsize{8}{10}\selectfont}
\usepackage{pifont}

\usepackage{colortbl}	

\makeatletter
\newcommand*{\rom}[1]{\expandafter\@slowromancap\romannumeral #1@}
\makeatother

\newcommand\ghat[2]{\hm{[} #2 \hm{]}^\wedge_{#1}}
\newcommand\gvee[2]{\hm{[} #2 \hm{]}^\vee_{#1}}

\newcommand\gexphat[1]{\text{exp}_{#1}^\wedge}
\newcommand\glogvee[1]{\text{log}_{#1}^\vee}
\newcommand\gadj[1]{\text{Ad}_{#1}}
\newcommand\gjac[1]{\Phi_{#1}}

\newcommand\kpred{_{k + 1 \mid k}}
\newcommand\kprev{_{k \mid k}}
\newcommand\knext{_{k+1 \mid k+1}}
\newcommand\kpone{_{k+1}}

\usepackage{xspace}

\newcommand{\DLGEKF}{DILIGENT-KIO}

\title{\LARGE \bf
DILIGENT-KIO: A Proprioceptive Base Estimator for Humanoid Robots using Extended Kalman Filtering on Matrix Lie Groups
}

\author{Prashanth Ramadoss$^{1, 2}$, Giulio Romualdi$^{1, 2}$, Stefano Dafarra$^{1}$, \\ Francisco Javier Andrade Chavez$^{3}$, Silvio Traversaro$^{4}$, and Daniele Pucci$^{1}$ 
\thanks{$^{1}$ Dynamic Interaction Control, Italian Institute of Technology,
Genoa, Italy, {\tt\small (e-mail: name.surname@iit.it)}}
\thanks{$^{2}$ DIBRIS, University of Genoa, Genoa, Italy}
\thanks{$^{3}$ Human Centered Robotics and Machine Intelligence, University of Waterloo, Ontario, Canada}
\thanks{$^{4}$ iCub Tech, Italian Institute of Technology,
Genoa, Italy}
}

\begin{document}

\maketitle
\thispagestyle{empty}
\pagestyle{empty}


\begin{abstract}
This paper presents a contact-aided inertial-kinematic floating base estimation for humanoid robots considering an evolution of the state and observations over matrix Lie groups.
This is achieved through the application of a geometrically meaningful estimator which is characterized by concentrated Gaussian distributions.
The configuration of a floating base system like a humanoid robot usually requires the knowledge of an additional six degrees of freedom which describes its base position-and-orientation. 
This quantity usually cannot be measured and needs to be estimated.
A matrix Lie group, encapsulating the position-and-orientation and linear velocity of the base link, feet positions-and-orientations and Inertial Measurement Units' biases, is used to represent the state while relative positions-and-orientations of contact feet from forward kinematics are used as observations. 
The proposed estimator exhibits fast convergence for large initialization errors owing to choice of uncertainty parametrization.
An experimental validation is done on the iCub humanoid platform.
\looseness=-1
\end{abstract}

\section{Introduction}
\par
State estimation remains an active research domain in the field of robotics.
In the case of Humanoid Robotics, the model uncertainties, the  unpredictable surrounding environment, and the often large number of robot sensors for both kinematic and dynamic quantities pose deep questions about the fundamental topic of sensor fusion. When attempting to answer these questions, the Lie group geometry of the underlying floating base system represents a further threat
for controllers aiming at robot stabilization.
This paper contributes towards the development of a proprioceptive base estimator for humanoid robots assuming that both the states and observations evolve over a space of distinct matrix Lie groups.

\par

In the legged robotics community, a common approach to achieve a \emph{reliable} proprioceptive base estimation is to fuse high frequency Inertial Measurement Units' (IMU) and kinematic sensors along with contact state information. 
The approach was first applied 
using
a strap-down IMU and a rigid body kinematics model, which can lead  to an Observability Constrained Extended Kalman Filter (OCEKF) to estimate the base \emph{position-and-orientation} of a quadrupedal robot with point foot contacts \cite{bloesch2013state}. 
These results 
can be
extended  to scenarios that  
incorporate the rotational constraints imposed by a humanoid robot's flat foot \cite{rotella2014state, piperakis2018nonlinear}.
Although inclusion of foot rotations within the state and observation models yielded better estimator performance, 
the approach seems to suffer from a lack of convergence guarantees in the case of large initialization errors. \looseness=-1

\par 
Alternately, the theory of invariant observer design is a solid framework to formulate the estimation problem using the theory of matrix Lie groups and exploit the symmetries of the underlying system \cite{barrau2016invariant, bonnabel2009non}. 
It can be applied, for instance, to derive a non-linear observer for robots with point foot contacts \cite{hartley2020contact}.
This estimation design method presents the interesting property that the evolution of the estimation error is independent from the current state estimate, thereby resulting in stronger convergence and consistency guarantees. 
A $SE_{N+2}$ matrix Lie group is defined to include the base \emph{position-and-orientation} and linear velocity along with the feet positions in the state representation by exploiting the property of semi-direct products of rotations.
Invariant measurement updates are exploited for contact feet relative position measurements evolving in Euclidean spaces. The property of invariant updates, however, cannot be directly used when feet rotations are introduced in the observations, so the approach cannot be applied on humanoid robots having flat feet. 
Further, including the feet rotations within the state violates the semi-direct product rule. So, the $SE_{N+2}(3)$ group may not be used anymore unless the feet positions-and-orientations are considered in a decoupled manner.

\par

In the context of  matrix Lie groups, this paper presents an estimator for the IMU biases and the humanoid robot floating base and feet \emph{positions-and-orientations}, hereafter referred to as \emph{poses}. 
The model observations consists of the full relative pose measurements obtained from robot forward kinematics.
In a sense, we extend the application scenarios of~\cite{hartley2020contact} to flat contact surfaces while retaining \emph{fast} estimation convergence-- see Table~\ref{table:soa}. 
More precisely, the contributions of the present paper follow.
\begin{itemize}
 	\item Development of a DIscrete LIe Group ExteNded kalman filTer for a contact-aided Kinematic-Inertial Odometry (DILIGENT-KIO) 
 	having
 	both the state and the observations evolving over distinct matrix Lie groups and investigating the benefits of the proposed modeling choice for the estimator design. \looseness=-1
 	\item Validation of the estimator through experiments using the  iCub humanoid platform, with a performance comparison with the state of the art.
\end{itemize}
We exploit the theory of EKF over matrix Lie groups using Concentrated Gaussian Distributions (CGD) \cite{bourmaud2013discrete, bourmaud2015continuous}. 
The uncertainty representation enabled by the matrix Lie group structure allows us to formulate an estimator with good convergence properties which improves over the existing work for humanoid robots.

\vspace{1mm}
\begin{table}[t]
\caption{Comparison with the State-of-the-art}
\centering
\scalebox{0.8}{
\begin{tabular}{p{1.8cm} | p{1.5cm} | p{1.5cm} | p{1cm}| p{0.8cm} }                                                    
\hline \rowcolor[gray]{.9}
Author, Year  &    State    &   Kinematic measurement     &   Support for flat contact surfaces    &    Fast Convergence        \\
\hline
Rotella, 2014 (OCEKF) \cite{rotella2014state}  &  Euclidean + Unit quaternion      &    Euclidean + Unit quaternion    &   \checkmark    &    \ding{55}      \\
\rowcolor[gray]{.9} Hartley, 2020 (InvEKF) \cite{hartley2020contact} &    Matrix Lie group    &   Euclidean     &  \ding{55}     &    \checkmark        \\
Ramadoss, 2021 (\DLGEKF{}) &   Matrix Lie group      &   Matrix Lie group     &   \checkmark    &            \checkmark \\
\hline

\end{tabular}
}
\label{table:soa} 
\vspace{-3mm}
\end{table}
\rowcolors{0}{}{}

This paper is organized as follows. Section \ref{sec:BACKGROUND} introduces the mathematical concepts for understanding the \DLGEKF{} algorithm. 
Section \ref{sec:ESTIMATION} describes the modeling of the system dynamics and observation models using matrix Lie groups and necessary Jacobian computations for the filter. 
This is followed by experimental evaluation of the proposed filter in Section \ref{sec:RESULTS} and concluding remarks in Section \ref{sec:CONCLUSION}. 

\section{Mathematical Background}
\label{sec:BACKGROUND}

\par

\subsection{Notations and definitions}
\subsubsection{Coordinate Systems}
\begin{itemize}
	
	\item $\mathcal{C[D]}$ denotes a frame with origin $o_\mathcal{C}$ and orientation $\mathcal{D}$;
	\item $\prescript{\mathcal{A}}{}{o}_\mathcal{B}\in \mathbb{R}^3$ and $\prescript{\mathcal{A}}{}{R}_\mathcal{B}\in SO(3)$ are the position and orientation of a frame $\mathcal{B}$ with respect to the frame $\mathcal{A}$;
	
	\item given $\prescript{\mathcal{A}}{}{o}_\mathcal{C}$ and $\prescript{\mathcal{B}}{}{o}_\mathcal{C}$,  $\prescript{\mathcal{A}}{}{o}_\mathcal{C} = \prescript{\mathcal{A}}{}{R}_\mathcal{B} \prescript{\mathcal{B}}{}{o}_\mathcal{C} + \prescript{\mathcal{A}}{}{o}_\mathcal{B}= \prescript{\mathcal{A}}{}{H}_\mathcal{B} \prescript{\mathcal{B}}{}{\bar{o}}_\mathcal{C}$, where $\prescript{\mathcal{A}}{}{H}_\mathcal{B} \in SE(3)$ is the homogeneous transformation and $\prescript{\mathcal{B}}{}{\bar{o}}_\mathcal{C} = (\prescript{\mathcal{B}}{}{o}_\mathcal{C}; \ 1) \in \mathbb{R}^4$ is the homogeneous representation of $\prescript{\mathcal{B}}{}{o}_\mathcal{C}$; 
	
	\item $\prescript{\mathcal{B[A]}}{}{v}_\mathcal{A, B} = \prescript{\mathcal{A}}{}{\dot{o}}_\mathcal{B} = \frac{d}{dt}(\prescript{\mathcal{A}}{}{o}_\mathcal{B}) \in \mathbb{R}^3$ denotes the linear part of a mixed-trivialized velocity \cite[Section 5.2]{traversaro2019multibody} between the frame $\mathcal{B}$ and the frame $\mathcal{A}$ expressed in the frame $\mathcal{B[A]}$. $\prescript{\mathcal{A}}{}{\omega}_\mathcal{A, B} \in \mathbb{R}^3$ denotes the angular velocity between the frame $\mathcal{B}$ and the frame $\mathcal{A}$ expressed in $\mathcal{A}$; \looseness=-1
	
	\item $\mathcal{A}$ denotes an absolute or an inertial frame;
	\item $\mathcal{B}$, $\mathcal{LF}$, $\mathcal{RF}$ and $\mathcal{S}$ indicate the frames attached to the base link, left foot, right foot and the IMU rigidly attached to the base link respectively;
\end{itemize}	
\subsubsection{Lie Groups}
For a gentle introduction to the theory of Lie groups applied to state estimation for robotics, we refer the readers to \cite[Chapter 7]{barfoot2017state}, \cite{sola2018micro, bourmaud2015continuous}. 
We will rely on the notation for Lie groups used by \cite{bourmaud2015continuous} in this paper.\looseness=-1

\begin{figure}[!t]
	\centering
	\includegraphics[scale=0.16]{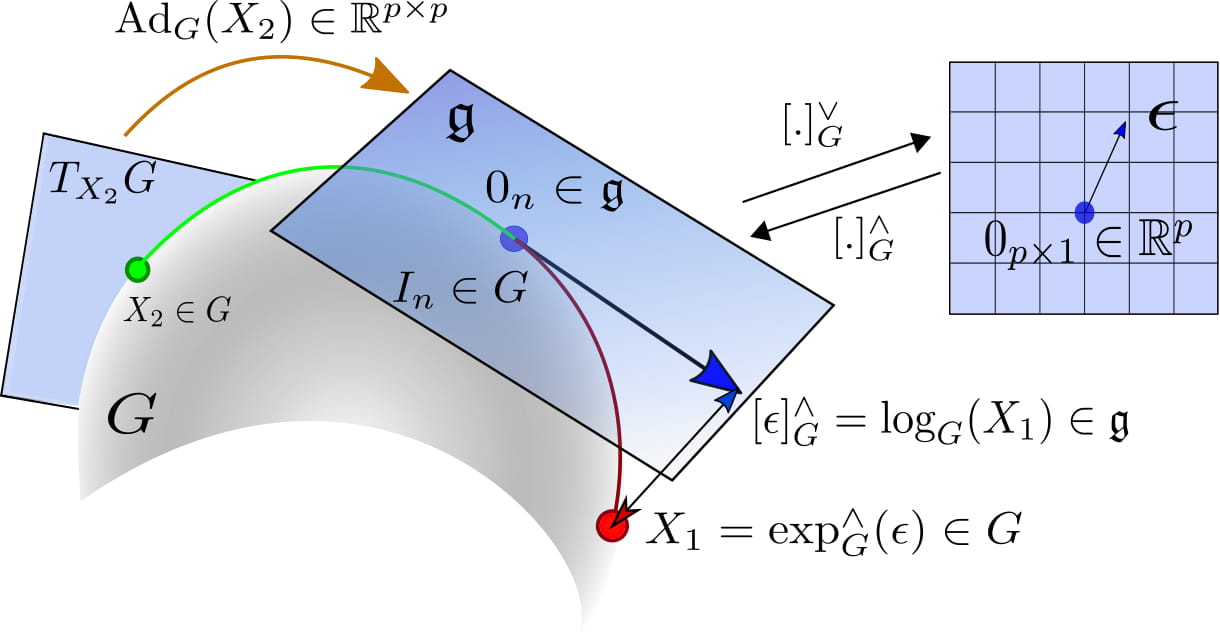}
	\caption{ An illustration of a Lie Group and its operators.  
\label{fig:liegroup}}
\vspace{-6mm}
\end{figure}

\begin{itemize}
\item $G, G^\prime \subset \mathbb{R}^{n \times n}$ denote matrix Lie groups and $X, Y \in G$ are elements of the matrix Lie group $G$. 
\item $\mathfrak{g, g^\prime} \subset \mathbb{R}^{n \times n}$ denote the matrix Lie algebras for the groups $G, G^\prime$ respectively.
\item $\ghat{G}{.}: \mathbb{R}^p \to \mathfrak{g}$ and $\gvee{G}{.}: \mathfrak{g} \to \mathbb{R}^p$ are the \emph{hat} and \emph{vee} operators for the matrix Lie group $G$ which denote a linear isomorphism between $\mathfrak{g}$ and a $p$-dimensional vector space. $\forall \;\mathfrak{a} \in \mathfrak{g}$, \; $a = \gvee{G}{\mathfrak{a}} \in \mathbb{R}^p$, $\mathfrak{a} = \ghat{G}{a}$.
\item $\gexphat{G}: \mathbb{R}^p \to G$ is the exponential map operator that maps elements from the vector space directly to elements of the group. $\forall a \in \mathbb{R}^p, \; \gexphat{G}(a) = \text{exp}(\ghat{G}{a})$.
\item $\glogvee{G}: G \to \mathbb{R}^p$ is the logarithm map operator that maps elements of the group directly to the vector space. $\forall \; X \in G, \; \glogvee{G}(X) = \text{log}(\gvee{G}{X})$. This mapping may not be unique. \looseness =-1
\item $\gadj{G}: \mathbb{R}^p \to \mathbb{R}^p$ is the adjoint matrix operator that linearly transforms vectors of the tangent space at an element X onto the Lie algebra which can be computed by, $\forall \; X \in G, a \in \mathbb{R}^p, \mathfrak{a} \in \mathfrak{g}, \;  \gadj{G}(X)\;a = \hm{[}X\;\mathfrak{a}\;X^{-1}\hm{]^\vee}_G$.  \looseness =-1
\item $\forall \; a \in \mathbb{R}^p, \; \gjac{G}(-a)$ denotes the right Jacobian of the matrix Lie group that relates any perturbations in the parametrizations of the Lie group to the changes in the group velocities $X^{-1} \dot{X}$. Similarly, the left Jacobian $\gjac{G}(a)$ relates those to the changes in $\dot{X}X^{-1}$ \cite{chirikjian2011stochastic}.
\end{itemize}
 A graphical representation of the Lie group operations is made in Figure \ref{fig:liegroup}.  \looseness=-1

\subsubsection{Miscellaneous}
\begin{itemize}
\item $I_n$ and $0_n$ denote the $n \times n$ identity and zero matrices;

\item given $w \in \mathbb{R}^3$ the \emph{hat operator} for $SO(3)$ is $S(.): \mathbb{R}^3 \to \mathfrak{so}(3)$, where $\mathfrak{so}(3)$ is the set of skew-symmetric matrices and $S(x) \; y = x \times y$.  $\times$ is the cross product operator in $\mathbb{R}^3$;

\item The following identity will be useful for our derivations, 
\begin{equation}
\small
\label{eq:SO3adjid}
\forall u \in \mathbb{R}^3, R \in SO(3), \; \; S(R u) = R S(u) R^T.
\end{equation}
\end{itemize}

\subsection{Matrix Lie Groups of interest}
\vspace{1mm}
\begin{table*}[!t]
	\caption{Matrix Lie groups depicting the group of rotations, rigid body transformation, double-direct spatial isometries \cite{barrau2016invariant}, and translations, respectively. }
	\label{table:matrixliegroups}
	\begin{center}
		\begin{tabular}{|c|c|c|c|c|c|c|c|}
			\hline
\rowcolor[gray]{.9}			Lie Group $G$ & Dim. $p$ & $X \in G$ & $\epsilon \in \mathbb{R}^p$ & $\ghat{G}{\epsilon} \in \mathfrak{g}$ & $X_1 \circ X_2 \in G$ & $X^{-1}  \in G$ & $X_{I} \in G$ \\
			\hline
			$SO(3)$  & $3$ & $ R \in \mathbb{R}^{3\times3}$ & $\omega \in \mathbb{R}^3$ & $S(\omega) \in \mathbb{R}^{3\times3}$ & $R_1 R_2$ & $R^T$ & $I_3$ \\
			\hline
		$SE(3)$  & $6$ & \scalebox{0.8}{$(p, R) = \begin{bmatrix}
			R & p \\ 0_{1 \times 3} & 1
			\end{bmatrix}  \in \mathbb{R}^{4\times4}$} &  \scalebox{0.8}{$\begin{bmatrix}
			v \\ \omega
			\end{bmatrix}  \in \mathbb{R}^6$} & \scalebox{0.8}{$\begin{bmatrix}
				S(\omega) & v \\ 0_{1 \times 3} & 0
			\end{bmatrix}$} & \scalebox{0.8}{$(R_1 p_2 + p_1, R_1 R_2)$} & \scalebox{0.8}{$(-R^T p, R^T)$} & \scalebox{0.8}{$(0_{3 \times 1}, I_3)$} \\
			\hline
			$SE_2(3)$  & $9$ & \scalebox{0.7}{$(p, R, v) = \begin{bmatrix}
				R & p & v \\ 0_{1 \times 3} & 1 & 0 \\ 0_{1 \times 3} & 0 & 1
				\end{bmatrix}  \in \mathbb{R}^{5\times5}$} &  \scalebox{0.7}{$\begin{bmatrix}
				v \\ \omega \\ a
				\end{bmatrix}  \in \mathbb{R}^9$} & \scalebox{0.7}{$\begin{bmatrix}
				S(\omega) & v & a \\ 0_{1 \times 3} & 0 & 0 \\  0_{1 \times 3} & 0 & 0
				\end{bmatrix}$} & \scalebox{0.7}{$(R_1 p_2 + p_1, R_1 R_2, R_1 v_2 + v_1)$} & \scalebox{0.7}{$(-R^T p, R^T, -R^T v)$} & \scalebox{0.7}{$(0_{3 \times 1}, I_3, 0_{3 \times 1})$} \\
		\hline
			$T(6)$  & $6$ & \scalebox{0.8}{$ b = \begin{bmatrix}
				I_6 & b \\ 0_{1 \times 6} & 1
				\end{bmatrix}  \in \mathbb{R}^{7\times7}$} &  \scalebox{0.8}{$b  \in \mathbb{R}^6$} & \scalebox{0.8}{$\begin{bmatrix}
				0_6 & b \\ 0_{1 \times 6} & 0
				\end{bmatrix}$} & \scalebox{0.8}{$ b_1 + b_2$} & \scalebox{0.8}{$-b$} & \scalebox{0.8}{$0_{6 \times 1}$} \\
			\hline
		\end{tabular}
	\end{center}
	\vspace{-6mm}
\end{table*}

The matrix Lie groups that we use to construct the estimation problem and their corresponding operators are described in the Table \ref{table:matrixliegroups}. 
Additionally, the property that the product of Lie groups is a Lie group \cite{bourmaud2015continuous, rudkovskii1997twisted} is used to build a composite state space and observation representation.
This is done through a combination of \emph{direct} and \emph{semi-direct} products. We will also exploit the property that a composite state space obtained through direct products result in non-interacting manifolds and corresponding non-interacting operators \cite[Section IV]{sola2018micro}. \looseness=-1

\subsection{Discrete Extended Kalman Filter on Matrix Lie Groups}

The concept of Concentrated Gaussian Distribution (CGD) is used to define the notion of uncertainty for the matrix Lie groups \cite{bourmaud2013discrete, bourmaud2015continuous, barfoot2017state, cesic2017mixture} producing a distribution on $G$ centered at $\hat{X}$,
\begin{equation}
\small
 X = \hat{X} \ \gexphat{G}(\epsilon) \quad \epsilon \sim \mathcal{N}_{\mathbb{R}^p}(0_{p, 1}, P),
\end{equation}
where, $\hat{X} \in G \subset \mathbb{R}^{n \times n}$ is the mean of $X$ and $\epsilon \in \mathbb{R}^p$ is a small perturbation having a zero-mean Gaussian distribution defined in the $p$-dimensional local vector space associated to the state, with the covariance matrix $P \in \mathbb{R}^{p \times p}$. It must be noted that for the CGD, the mean is defined on the group and the covariance is defined in the Lie algebra.
This remains a reasonable representation of uncertainty over Lie groups if the maximum eigenvalues of $P$ are sufficiently small. \looseness=-1

Given a discrete dynamical system on matrix Lie groups, \looseness=-1
\vspace{-0.3em}
\begin{align}
\small
\label{eq:liegroupsystem}
X_{k+1} &=  X_k \; \text{exp}_{G}^{\wedge}\big(\Omega(X_k, u_k) + w_k\big) \\
\label{eq:liegroupmeas}
z_k &= h (X_k) \; \text{exp}^{\wedge}_{G^\prime}\big(n_k\big),
\end{align}
\vspace{-0.3em}
$\Omega : G \times \mathbb{R}^m \to \mathbb{R}^p$ is the left trivialized velocity of the matrix Lie group expressed as a function of the state $X_k \in G$ and an exogenous control input $u_k \in \mathbb{R}^m$ at an instant $k$.
$w_k\sim \mathcal{N}_{\mathbb{R}^p}(0_{p,1},  Q_k)$ is a Gaussian white noise with covariance $Q_k$ in $\mathbb{R}^p$. \looseness=-1

The observations $z_k$ are considered to be evolving over a matrix Lie group $G^\prime$ of dimensions $l$  distinct from the state space $G$. $h: G \to G^\prime$ is the measurement model mapping the states $X \in G$ to the space of observations $G^\prime$. $n_k\sim\mathcal{N}_{\mathbb{R}^{q}}(0_{q,1}, \; N_k)$ is the measurement noise described as a Gaussian white noise with covariance $N_k$ defined in the $q$-dimensional vector space of the observations.\looseness=-1

The Discrete EKF on matrix Lie groups, using the left invariant error formulation $\hat{X}^{-1} X$, is described in Algorithm \ref{algo:dlgekf}. 
The algorithmic structure reduces to the regular EKF algorithm if we consider $G$ and $G^\prime$ as Euclidean spaces. \looseness=-1

\scalebox{0.87}{\parbox{.5\linewidth}{
\begin{algorithm}[H]
\small
  \caption{Discrete EKF on matrix Lie groups \cite{bourmaud2013discrete}}\label{algo:dlgekf}
  \begin{algorithmic}
    \State \textbf{Input:} $\hat{X}\kprev, P\kprev, z\kpone, u_k$
    \State \textbf{Output:} $\hat{X}\knext, P\knext$
    \\\hrulefill
    \State \textbf{Propagation:}\\
    $\hat{X}\kpred = \hat{X}\kprev \; \text{exp}_G^\wedge\big(\Omega(\hat{X}\kprev, u_k)\big)$ \\
    $P\kpred = F_k P\kprev F_k^T \; + $ \\ $\qquad \; \Phi_G\big(-\Omega(\hat{X}\kprev, u_k)\big)\,Q_k\,\Phi_G\big(-\Omega(\hat{X}\kprev, u_k)\big)^T$ \\
    \hrulefill
    \State \textbf{Update:} \\
    $K\kpone = P\kpred\;H\kpone^T\;(H\kpone\;P\kpred\;H\kpone^T +\; N\kpone)^{-1}$ \\ 
    $\tilde{z}\kpone = \text{log}_{G^\prime}^\vee\big(h(\hat{X}\kpred)^{-1}\;z\kpone\big)$ \\
    $m^{-}\kpone = K\kpone \tilde{z}\kpone$ \\
    $\hat{X}\knext =  \hat{X}\kpred \; \text{exp}^{\wedge}_G\big(m^{-}\kpone\big)$ \\
    $P\knext = \Phi_G\big(-m^{-}\kpone\big)\;(I_p - K\kpone\;H\kpone)P\kpred\;\Phi_G\big(-m^{-}\kpone\big)^T$ \\
    \hrulefill 
    \State where, \\
    $F_k = \text{Ad}_G\big(\text{exp}^{\wedge}_G(-\Omega(\hat{X}\kprev, u_k))\big) + \Phi_G\big(-\Omega(\hat{X}\kprev, u_k)\big)\mathcal{F}_k$  \\ 
    \vspace{2mm}
    $\mathcal{F}_k = \frac{\partial}{\partial \epsilon}\Omega\big(\hat{X}\kprev\;\text{exp}^{\wedge}_G(\epsilon)\big)\big|_{\epsilon = 0}$\\ \vspace{2mm}
    $H_k = \frac{\partial}{\partial \epsilon}\;\text{log}^{\vee}_{G^\prime}\big(h^{-1}\big(\hat{X}\kpred\big)\;h\big(\hat{X}\kpred\;\text{exp}^{\wedge}_G(\epsilon)\big)\big)\big|_{\epsilon = 0}$\\
    \vspace{-1mm}
  \end{algorithmic}
  \vspace{-2mm}
\end{algorithm}
}}


\section{DILIGENT-KIO: DIscrete LIe Group ExteNded kalman filTer for Kinematic-Inertial Odometry}
\label{sec:ESTIMATION}

\subsection{State representation}
We wish to estimate the position $\prescript{\mathcal{A}}{}{o}_\mathcal{B}$, orientation $\prescript{\mathcal{A}}{}{R}_\mathcal{B}$ and linear velocity of the base link $\prescript{\mathcal{A}}{}{\dot{o}}_\mathcal{B}$ in an inertial frame. 
Additionally we also consider the feet positions $\prescript{\mathcal{A}}{}{o}_\mathcal{F}$ and orientations $\prescript{\mathcal{A}}{}{R}_\mathcal{F}$, where $\mathcal{F} = \{ \mathcal{LF}, \mathcal{RF} \}$ in the state. 
Further, to extend the implementation for the real-world hardware applications, it is necessary to estimate the slowly time-varying biases $(b_a, b_g)$ affecting the accelerometer and gyroscope measurements from the IMU. It must be noted that the biases are always expressed in the local IMU frame.

For the sake of readability, we will use the following short-hand notations in the rest of the paper.
We will denote the tuple representing the base link quantities $ X_\text{base} = (\prescript{\mathcal{A}}{}{o}_\mathcal{B}, \prescript{\mathcal{A}}{}{R}_\mathcal{B}, \prescript{\mathcal{A}}{}{\dot{o}}_\mathcal{B})$ as $(p, R, v)$, while the feet quantities $X_f = (\prescript{\mathcal{A}}{}{o}_\mathcal{F}, \prescript{\mathcal{A}}{}{R}_\mathcal{F})$ are denoted as $(d_f, Z_f)$, where $f = \{l, r\}$. 
The biases $(b_a, b_g)$ are sometimes included in a single vector as  $b$. \looseness=-1

As seen earlier, due to the introduction of the feet rotations in the state space, the $SE_{N+2}(3)$ Lie group cannot be used to model the state representation. 
Instead, we use a non-interacting composite matrix Lie group $SE_2(3) \times SE(3)^2 \times T(6) \subset \mathbb{R}^{20 \times 20}$  to represent the state space $\mathcal{M}$,

\scalebox{0.75}{\parbox{.5\linewidth}{
\setcounter{MaxMatrixCols}{20}
\begin{align*}
\footnotesize
X &=
\begin{bmatrix}
R & p & v & 0_3  & 0_{3,1} & 0_3 & 0_{3,1} & 0_{3,6} &  0_{3,1} \\
0_{1,3} & 1 & 0 & 0_{1,3} & 0  & 0_{1,3}  & 0  & 0_{1,6}  &  0  \\
0_{1,3} & 0 & 1 & 0_{1,3} & 0  & 0_{1,3}  & 0  & 0_{1,6}  &  0   \\
0_3 & 0_{3,1}  & 0_{3,1}  & Z_l & d_l & 0_3  & 0_{3,1} & 0_{3,6} & 0_{3,1}  \\
0_{1,3} & 0 & 0 & 0_{1,3} & 1  & 0_{1,3}  & 0  & 0_{1,6}  &  0 \\
0_3 & 0_{3,1} & 0_{3,1}  & 0_3  & 0_{3,1}  & Z_r & d_r &0_{3,6} & 0_{3,1} \\
0_{1,3} & 0 & 0 & 0_{1,3} & 0  & 0_{1,3}  & 1  & 0_{1,6}  &  0 \\
0_{6,3} & 0_{6,1}  &  0_{6,1} & 0_{6,3} &  0_{6,1} &  0_{6,3} & 0_{6,1}  & I_{6} & b  \\
0_{1,3} & 0 & 0 & 0_{1,3} & 0  & 0_{1,3}  & 0  & 0_{1,6}  &  1 
\end{bmatrix} \in \mathcal{M}.
\end{align*}
}}

The hat operator $\ghat{\mathcal{M}}{\epsilon}$ mapping the vectors $\epsilon \in \mathbb{R}^{27}$ to the Lie algebra $\mathfrak{m}$ becomes, 

\scalebox{0.7}{\parbox{.5\linewidth}{
		\begin{align*}
		\small
			\ghat{\mathcal{M}}{\epsilon} &=
			\begin{bmatrix}
			S(\epsilon_R) & \epsilon_p & \epsilon_v & 0_3  & 0_{3,1} & 0_3 & 0_{3,1} & 0_{3,6} &  0_{3,1} \\
			0_{1,3} & 0 & 0 & 0_{1,3} & 0  & 0_{1,3}  & 0  & 0_{1,6}  &  0  \\
			0_{1,3} & 0 & 0 & 0_{1,3} & 0  & 0_{1,3}  & 0  & 0_{1,6}  &  0   \\
			0_3 & 0_{3,1}  & 0_{3,1}  & S(\epsilon_{Z_l}) & \epsilon_{d_l} & 0_3  & 0_{3,1} & 0_{3,6} & 0_{3,1}  \\
			0_{1,3} & 0 & 0 & 0_{1,3} & 0  & 0_{1,3}  & 0  & 0_{1,6}  &  0 \\
			0_3 & 0_{3,1} & 0_{3,1}  & 0_3  & 0_{3,1}  & S(\epsilon_{Z_r}) & \epsilon_{d_r} &0_{3,6} & 0_{3,1} \\
			0_{1,3} & 0 & 0 & 0_{1,3} & 0  & 0_{1,3}  & 0  & 0_{1,6}  &  0 \\
			0_{6,3} & 0_{6,1}  &  0_{6,1} & 0_{6,3} &  0_{6,1} &  0_{6,3} & 0_{6,1}  & 0_{6} & \epsilon_{b}  \\
			0_{1,3} & 0 & 0 & 0_{1,3} & 0  & 0_{1,3}  & 0  & 0_{1,6}  &  0 
			\end{bmatrix} \in \mathfrak{m}
		\end{align*}
}}

with the vector $\epsilon \in \mathbb{R}^{27}$ as,
\vspace{-0.4em}
\begin{align}
\small
\begin{split}
\label{eq:statevec}
\epsilon_\mathcal{M} &= \begin{bmatrix}\epsilon^{T}_{p} & \epsilon^{T}_{R}& \epsilon^{T}_{v} & \epsilon^{T}_{d_l} & \epsilon^{T}_{Z_l} 
& \epsilon^{T}_{d_r} & \epsilon^{T}_{Z_r} & \epsilon^{T}_{b}
\end{bmatrix}^{T}.
\end{split}
\end{align}
\vspace{-0.4em}
The exponential mapping is given by,
\begin{equation}
\resizebox{.47\textwidth}{!}{%
  $\begin{aligned}
  \label{eq:expmap}
&\gexphat{\mathcal{M}}(\epsilon) = \big(J(\epsilon_{R})\; \epsilon_{p}, \; \; \gexphat{SO(3)}(\epsilon_{R}), \; \; J(\epsilon_{R})\; \epsilon_{v}, \; \; J(\epsilon_{Z_l})\; \epsilon_{d_l}, \\
& \quad \quad \quad \quad \gexphat{SO(3)}(\epsilon_{Z_l}),  \; \; J(\epsilon_{Z_r})\; \epsilon_{d_r}, \; \; \gexphat{SO(3)}(\epsilon_{Z_r}),
\; \epsilon_{b}\big).
\end{aligned}$%
}
\end{equation}

where, $J$ and $\gexphat{SO(3)}$ are the left Jacobian and the exponential map of $SO(3)$ (see \cite{barfoot2014associating} for closed-form expressions). \looseness=-1
\vspace{-0.2em}
The adjoint matrix operator is given as,
\begin{align}
\small
\begin{split}
\gadj{\mathcal{M}}(X) &= \text{blkdiag} \big( \gadj{SE_2(3)}(X_\text{base}), \; \gadj{SE(3)}(X_l), \\ &  \quad \quad \quad \qquad \gadj{SE(3)}(X_r), \; I_6 \big).
\label{eq:adjmap}
\end{split}
\end{align}

The left Jacobian of the matrix Lie group is given as,
\begin{align}
\small
\begin{split}
\gjac{\mathcal{M}}(\epsilon) &= \text{blkdiag} \big( \gjac{SE_2(3)}(\epsilon_\text{base}), \; \gjac{SE(3)}(\epsilon_{l}), \\ &  \quad \quad \quad \qquad \gjac{SE(3)}(\epsilon_{r}), \; I_6 \big).
\label{eq:ljac}
\end{split}
\end{align}

The relevant matrix Lie group operator definitions for \eqref{eq:adjmap}, \eqref{eq:ljac} are provided in the Appendix.

\subsection{Discrete System Dynamics}
We use the discretized system dynamics similar to \cite{bloesch2013state, rotella2014state, hartley2020contact, piperakis2018nonlinear} in this work.
The evolution of the base pose is described by a rigid body kinematics model based on a strap-down IMU rigidly attached to the base link of the robot.
For simplicity, the coordinate frames of IMU ($\mathcal{S}$) and the base link ($\mathcal{B}$) are assumed to coincide. 
Without such an assumption, we need to be careful about the necessary coordinate transformations required to express quantities of $\mathcal{S}$ in $\mathcal{B}$.
The accelerometer measurements $\tilde{\alpha} _{\mathcal{A ,B}}^{g, \text{lin}}$ and gyroscope measurements $\prescript{\mathcal{B}}{}{\tilde{\omega}}_{\mathcal{A, B}}$ are passed as exogenous inputs $u_k$ to the system. These measurements are modeled to be affected by slowly time-varying biases, $b_a$ and $b_g$, and additive white Gaussian noise $\text{w}^\mathcal{B}_a$ and $\text{w}^\mathcal{B}_\omega$, respectively. Further, the biases are assumed to be affected by noise $\text{w}^\mathcal{B}_{b_a}$ and $\text{w}^\mathcal{B}_{b_g}$. \looseness=-1
\vspace{-0.25em}
\begin{equation}
\small
\label{eq:imumodel}
\begin{split}
\tilde{\alpha} _{\mathcal{A ,B}}^{g, \text{lin}}\,_{k} &= {R}_k^{T} ( \prescript{\mathcal{A}}{}{\ddot{o}}_{\mathcal{B}_k} - \prescript{\mathcal{A}}{}g) \;+ \; {b_a}\,_{k} \;+\; \text{w}^\mathcal{B}_a \\
\prescript{\mathcal{B}}{}{\tilde{\omega}}_{\mathcal{A, B}_k} &= \prescript{\mathcal{B}}{}{\omega}_{\mathcal{A, B}_k} \; + \; {b_g}_{k} \; + \; \text{w}^\mathcal{B}_\omega.
\end{split}
\end{equation}

A constant motion model is considered for the feet pose assuming the holonomic constraints imposed by the feet when making a rigid contact with the environment. Thus, the feet velocities are considered equal to zero and affected only by white Gaussian noise. These noises can be characterized as translational or rotational slippage of the feet, when expressed in the inertial frame \cite{rotella2014state}. 
The bias dynamics are also considered as constant and affected only by noise.

The discrete system dynamics is described as,
\vspace{-2em}
\begin{center}
	\begin{equation}
	\small
	\label{eq:dekfsystemdynamics}
	\begin{split}
	{p}_{k+1} =& \;{p}_{k} + \;  {v}_{k} \; \Delta T \; +  \frac{1}{2} \; {R}_{k}  \;\prescript{\mathcal{B}}{}{\alpha}_{\mathcal{A, B}_k} \; \Delta T^2 \\
	{R}_{k+1} =& \; {R}_{k} \; \gexphat{SO(3)}\big( \prescript{\mathcal{B}}{}{\omega}_{\mathcal{A, B}_k} \; \Delta T \big) \\
	{v}_{k+1} =& \; {v}_{k} + \; {R}_{k} \;\prescript{\mathcal{B}}{}{\alpha}_{\mathcal{A, B}_k}  \; \;  \Delta T \\
	{d_f}\,_{k+1} =& \; {d_f}\,_{k}  + \;{Z_f}\,_{k}\; \text{w}^\mathcal{F}_v \; \Delta T \\
	{Z_f}\,_{k+1} =& \; {Z_f}\,_{k} \; \gexphat{SO(3)}\big(\text{w}^\mathcal{F}_\omega \; \Delta T\big) \\
	{b}\,_{k+1} =& \; {b}\,_{k} +\; \text{w}^\mathcal{B}_{b} \; \Delta T
	\end{split}
	\end{equation}
\end{center}

where, $\small \prescript{\mathcal{B}}{}{\alpha}_{\mathcal{A, B}_k} = \big(\tilde{\alpha} _{\mathcal{A ,B}}^{g, \text{lin}}\,_{k}  - \; {b_a}\,_{k} - \; \text{w}^\mathcal{B}_a\big) + \; {R}_k^{T} \prescript{\mathcal{A}}{}g$ is the linear acceleration of the base with respect to the inertial frame expressed in the base frame. $\prescript{\mathcal{A}}{}g$ is the acceleration due to gravity expressed in the inertial frame. $\prescript{\mathcal{B}}{}{\omega}_{\mathcal{A, B}_k} = \; \prescript{\mathcal{B}}{}{\tilde{\omega}}_{\mathcal{A, B}_k} - \; {b_g} _{k}  - \; \text{w}^\mathcal{B}_\omega $ is the unbiased angular velocity expressed in the base frame. $\Delta T$ is the sampling period with which a zero-order hold on the inputs are assumed in order to discretize the continuous system dynamics. $\text{w}^\mathcal{F}_v$ and $\text{w}^\mathcal{F}_\omega$ are the white Gaussian noise affecting the null feet velocities, both expressed in the feet frames. 

Given Eqns. \eqref{eq:liegroupsystem} and \eqref{eq:dekfsystemdynamics}, the left trivialized motion model and the noise vector become,
\vspace{-0.5mm}
\begin{align}
\begin{split}
\label{eq:leftrivvel}
&\footnotesize{\Omega({X}, u) = \begin{bmatrix}
{R ^T}{v} \Delta T + \;\;\frac{1}{2}\prescript{\mathcal{B}}{}{\alpha}_{\mathcal{A, B}}\;\Delta T^2 \\
\prescript{\mathcal{B}}{}{\omega}_{\mathcal{A, B}} \Delta T \\ 
\prescript{\mathcal{B}}{}{\alpha}_{\mathcal{A, B}}\;\Delta T  \\ 
0_{18,1}
\end{bmatrix} \in \mathbb{R}^{27}},
\end{split}
\end{align}

\begin{equation}
\resizebox{.47\textwidth}{!}{%
  $\begin{aligned}
  \label{eq:leftrivnoise}
  & \text{w}  \footnotesize{ \;= \big(-0.5 \; \text{w} ^\mathcal{B}_a \Delta T, \; -\text{w} ^\mathcal{B}_\omega, \; -\text{w} ^\mathcal{B}_a, \; \text{w}^\mathcal{LF}_v, \text{w}^\mathcal{LF}_\omega, \; \text{w}^\mathcal{RF}_v, \text{w}^\mathcal{RF}_\omega, \; \text{w}^\mathcal{B}_{b} \big) \; \Delta T} \\
&\footnotesize{ \quad \qquad \qquad \qquad Q_k = \text{Cov}(\text{w}) \in \mathbb{R}^{27 \times 27}}.
\end{aligned}$%
}
\end{equation}

The state is propagated through the left trivialized motion model as $\hat{X}\kpred = \hat{X}\kprev \; \gexphat{\mathcal{M}}\big(\hat{\Omega}\big)$ using the current state estimate ${\hat{X}\kprev}$ and the IMU measurements $u_k$. 
The covariance propagation requires the computation of Jacobian of the left trivialized motion model at the current state estimate with an infinitesimal additive perturbation in the vector space, $ \Omega \big(\hat{X} \; \gexphat{\mathcal{M}}({\epsilon}_\mathcal{M}) \big)$. \looseness=-1

Considering Eqn. \eqref{eq:expmap} and assuming a first-order approximation for the exponential map and the left Jacobian of $SO(3)$ \cite{barfoot2014associating}, and neglecting cross-products of infinitesimal perturbations,  we have the tuple, \looseness=-1
\vspace{-0.5mm}
\begin{equation}
\resizebox{.47\textwidth}{!}{%
  $\begin{aligned}
  \label{eq:perturbedmean}
&\hat{X} \;\gexphat{\mathcal{M}}({\epsilon}) \approx \big( {R} \ {\epsilon}_{p} + {p}, \; \; {R} + {R} \ S({\epsilon}_{R}), \; \; {R} \ {\epsilon}_{v} + {v}, \; \; {Z_l} \ {\epsilon}_{d_l} + {d_l}, \\
& \quad \quad {Z_l} + {Z_l} \ S({\epsilon}_{Z_l}),  \; \; {Z_r} \ {\epsilon}_{d_r} + {d_r}, \; \; {Z_r} + {Z_r} \ S({\epsilon}_{Z_r}),
\; {b} + {\epsilon}_{b}\big).
\end{aligned}$%
}
\end{equation}

Subsequently, by substituting \eqref{eq:perturbedmean} in \eqref{eq:leftrivvel}, the Jacobian of the left trivialized motion model $\mathcal{F}_k$ can be computed in a straight-forward manner as,
\begin{equation}
\label{eq:leftparamvelJacobian}
\begin{split}
\small
&\mathcal{F}_k = \frac{\partial}{\partial \epsilon}\Omega\big(\hat{X}\;\text{exp}^{\wedge}_G(\epsilon)\big)\bigg|_{\epsilon = 0} = \\
&\begingroup
\setlength\arraycolsep{1.8pt}\footnotesize{\begin{bmatrix}
	0_3 & S(\Xi) & I_3 \; \Delta T & 0_{3, 12} &  -\frac{1}{2}I_{3}\;\Delta T^2 & 0_3 \\
	0_3 & 0_3 & 0_3 & 0_{3, 12} &  0_3 & -I_{3}\;\Delta T \\
	0_3 & S(\tilde{g}) & 0_3 & 0_{3, 12} &  -I_{3}\;\Delta T & 0_3 \\
	0_{18, 3} & 0_{18, 3} & 0_{18, 3} & 0_{18, 12} &  0_{18, 3} & 0_{18, 3}
	\end{bmatrix} }\endgroup\\
\end{split}
\end{equation}
where, $ \Xi = {R^T v} \Delta T + \frac{1}{2}\; {R^T} \prescript{\mathcal{A}}{}g \; \Delta T^2 $ and $\small \tilde{g} = {R^T} \prescript{\mathcal{A}}{}g \; \Delta T$. 

The contact states are assumed to be known and passed as inputs to the system. 
When a foot is not in contact, the variances related to the foot velocities are dynamically scaled to very high values causing the estimated foot pose from the prediction model to grow uncertain.
When a foot gains contact, the measurement updates cause the foot pose to reset to the new and correct estimate.
The absence of contacts is not handled within this formulation. \looseness=-1

\vspace{-0.3em}
\subsection{Forward Kinematics Measurement Model}
The measurement model is the relative pose between the frames of the feet in contact with the environment and the base link using the Forward Kinematics (FK).

The encoder measurements $\small \tilde{s} = s + \text{w}^s$ provide the joint positions $s$ affected by additive white Gaussian noise $\text{w}_s$. The relative pose of the feet with respect to the base link is computed using forward kinematics $\small \text{FK}(s): \mathbb{R}^{\text{DOF}} \to SE(3)$ as, \looseness=-1
\begin{equation}
\small
\begin{split}
\label{eq:fkin}
&\footnotesize{z^\mathcal{F}_{\text{SS}} = \prescript{\mathcal{B}}{}{H}_\mathcal{F} =  \text{FK}(\tilde{s})} \\
&\footnotesize{z_{\text{DS}} = \begin{bmatrix}
\prescript{\mathcal{B}}{}{H}_\mathcal{LF} & 0_4 \\ 0_4 & \prescript{\mathcal{B}}{}{H}_\mathcal{RF}
\end{bmatrix} = \begin{bmatrix}
\text{FK}(\tilde{s}) & 0_4 \\ 0_4 & \text{FK}(\tilde{s})
\end{bmatrix}}.
\end{split}
\end{equation}

The measurements $z^\mathcal{F}_{\text{SS}}$ evolve over $SE(3)$ during the single support and $z_{\text{DS}}$ over $SE(3)^2$ during double support. For the rest of this section, we will only focus on the single support scenario, since the same computations apply also for double support due to the rule of non-interacting manifolds as a result of the direct product.

As a function of the state, forward kinematic measurements can be written in the form of Eqn. \eqref{eq:liegroupmeas} as, 
\vspace{-0.2em}
\begin{equation}
\label{eq:measurementSS}
\footnotesize{z^\mathcal{F}_{\text{SS}} =  \begin{bmatrix} {R^T}{Z_f} & {R^T}({d_f}\;-\;{p})   \\
0_{1, 3} & 1
 \end{bmatrix} \; \gexphat{SE(3)}\left( \begin{bmatrix} \text{n}^\mathcal{F}_v \\ \text{n}^\mathcal{F}_\omega \end{bmatrix}\right)}.
\end{equation}
The left-trivialized forward kinematic noise is related to the encoder noise $\text{w}_s$ through the manipulator Jacobian $\prescript{\mathcal{F}}{}{J}_{\mathcal{B, F}}$ as, \looseness=-1
$$ \footnotesize{\text{n}^\mathcal{F} = \begin{bmatrix} \text{n}_v^\mathcal{F} \\ \text{n}_\omega^\mathcal{F}  \end{bmatrix} = \prescript{\mathcal{F}}{}{J}_{\mathcal{B, F}} \; \text{w}_s \in \mathbb{R}^6}, \; \; N\kpone = \text{Cov}(\text{n}^\mathcal{F}) \in \mathbb{R}^{6 \times 6}.$$

In order to update the state estimates using the FK measurements, the innovation $\tilde{z}\kpone$ is computed using the logarithm mapping of $SE(3)$ (see \cite[Appendix D]{sola2018micro} for definition). 
Further, the only required derivation is the measurement model Jacobian,

\scalebox{0.87}{\parbox{.5\linewidth}{
$$ H\kpone^\mathcal{F} = 
\frac{\partial}{\partial \epsilon}\;\glogvee{SE(3)}\big(h^{-1}\big(\hat{X}\kpred\big)\;h\big(\hat{X}\kpred\;\gexphat{\mathcal{M}}({\epsilon}_\mathcal{M})\big)\big)\bigg|_{\epsilon = 0}. $$
}}

Computing $\displaystyle h^{-1}( \hat{X} ) \ h(\hat{X} \; \gexphat{\mathcal{M}}({\epsilon}_\mathcal{M}) ) \ $ using the first order approximation of \eqref{eq:perturbedmean} as,

\scalebox{0.87}{\parbox{.5\linewidth}{
$$
\footnotesize{
\begin{bmatrix} 
I_3 + S({\epsilon}_{Z_f})- S({Z_f^T}\;{R}\;{\epsilon}_{R}) &  {\epsilon}_{d_f} - {Z_f^T}\;{R}\;{\epsilon}_{p} + {Z_f^T}\; S({R}\;{\epsilon}_{R})\; ({p - d_f}) \\
0_{1, 3} & 1
\end{bmatrix}.
}
$$
}}

Approximating the matrix logarithm assuming the perturbations are small, $\forall B \in \mathbb{R}^{l \times l}, \; \text{log}( B) \ \approx B-I_l\ $, and using the identities $S(x) y = -S(y) x$ and \eqref{eq:SO3adjid}, $\glogvee{SE(3)}\left(h^{-1}( \hat{X} ) \ h(\hat{X} \; \gexphat{\mathcal{M}}({\epsilon}_\mathcal{M}) )\right)$ becomes,
\vspace{-0.2em}
$$
\centering
\footnotesize{
	\begin{bmatrix} 
 {\epsilon}_{d_f} - \; {Z_f^T}{R}\;{\epsilon}_{p} - \; S({R^T(p-d_f)})\;{Z_f^TR}\;{\epsilon}_{R}\\
 {\epsilon}_{Z_f}- {Z_f^T}{R}\;{\epsilon}_{R}
	\end{bmatrix}.
}
$$
\vspace{-0.2em}
Finally, on further simplifications using the identities, and taking partial derivatives, the measurement model Jacobian during single support can be written as,
\vspace{-0.18em}
\begin{equation}
\small
\begin{split}
\label{eq:measmodeljacobianLF}
\footnotesize{
H^\mathcal{LF} = \begin{bmatrix} 
-{Z_l^T R} & -{Z_l^T}S({p - d_l}){R} & 0_3 & I_3 & 0_3 & 0_{3, 12} \\
0_3 & -{Z_l^T R}  & 0_3 & 0_3 & I_3 & 0_{3, 12} 
\end{bmatrix}}, \\
\footnotesize{
H^\mathcal{RF} = \begin{bmatrix} 
-{Z_r^T R} & -{Z_r^T}S({p - d_r}){R} & 0_{3, 9} & I_3 & 0_3 & 0_{3, 6} \\
0_3 & -{Z_r^T R}  & 0_{3, 9} & 0_3 & I_3 & 0_{3, 6} 
\end{bmatrix}}.
\end{split}
\end{equation}
\vspace{-0.2em}
During double support, these matrices are simply stacked together to form a composite Jacobian.
We now have all the necessary elements for the computation of Kalman gain $K\kpone$ for the measurement update and subsequently the state reparametrization using $m^{-}\kpone$. 
The presented method can similarly be extended for an arbitrary number of contacts.


\section{Experiments and Results}
\label{sec:RESULTS}
\begin{figure*}[t!]
	\begin{subfigure}[b]{\textwidth}
		\centering	
\includegraphics[width=\textwidth]{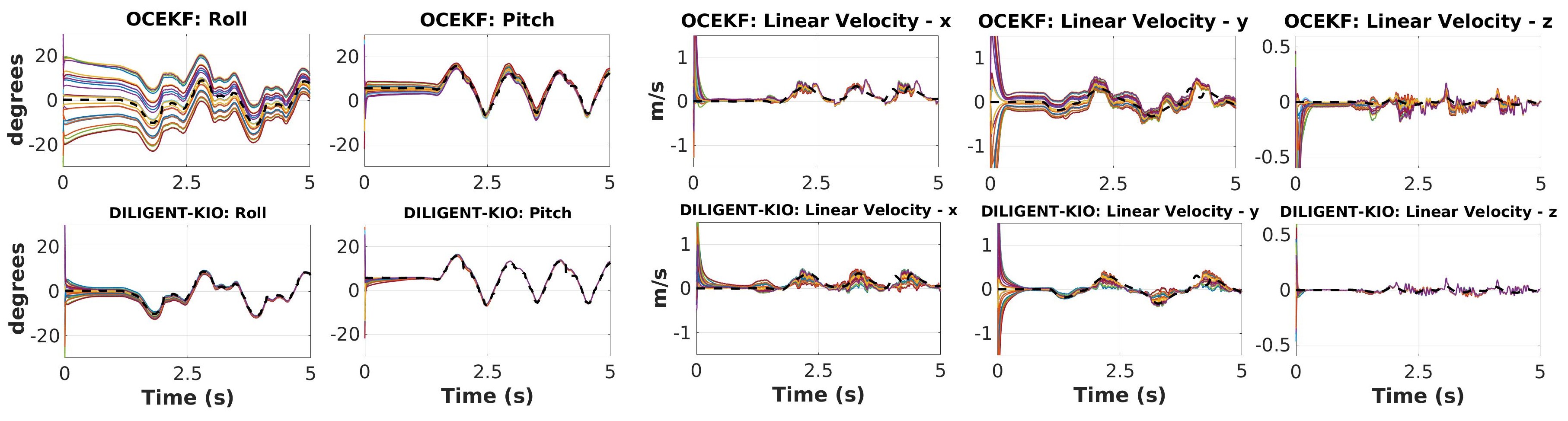}		
	\end{subfigure}
	\caption{\label{fig:random_initialstates_real} Orientation and velocity estimates from 25 trials of OCEKF and \DLGEKF{} (proposed) for a forward walking experiment on the iCub humanoid platform using the same noisy measurements from the robotic hardware, noise statistics and initial covariances, but initialized with random orientations and velocities. The dashed black line is the ground truth trajectory from the Vicon motion capture system. \DLGEKF{} (bottom row) is seen to converge considerably faster than OCEKF (top row) in almost all the directions. }
	\vspace{-2mm}
\end{figure*}
In this section, we present the results from the experimental evaluation of \DLGEKF{} on the iCub humanoid platform. \looseness=-1

\subsection{Experimental Setup}

A subset of 32 degrees-of-freedom (legs, torso, arms and neck) on the iCub that are equipped with joint encoders is used to measure the joint angles at $1000 \si{Hz}$. 
The robot is equipped with an XSens MTi-300 series IMU mounted in its base link providing linear accelerometer and gyroscope measurements at $100 \si{Hz}$.  
The contact states are inferred through a Schmitt trigger thresholding of contact wrenches estimated by the estimation algorithm presented in \cite{nori2015icubWBC} available at $100 \si{Hz}$.
The base estimation is run at $100 \si{Hz}$, and the encoder measurements are sub-sampled to the same frequency as the estimator. The estimators are validated on the robotic hardware with pose measurements from the Vicon motion capture system. \looseness=-1 
\vspace{-0.2em}

\subsection{Baseline algorithms for comparison}
We use the state-of-the-art filters,  Observability Constrained quaternion based EKF (OCEKF) described in \cite{rotella2014state} and Right Invariant EKF (InvEKF) described in \cite{hartley2020contact}, for a baseline comparison. 
It must be noted that while the former supports flat foot contacts, the latter only point foot contacts.

Further, we also use a Simple Weighted Averaging method (SWA) similar to \cite[Method I]{flayols2017experimental} as one of the baseline algorithms. 
In SWA, Legged Odometry (LO) is used to obtain the base pose estimates. 
The rotation estimated by LO is then fused with a rotation estimated by an off-the-shelf quaternion EKF. 
This fusion is done through a weighted rotation averaging over $SO(3)$ using Manton's convergent algorithm \cite[Section ~5.1]{hartley2013rotation}, which computes the average in the tangent space and projects it back to $SO(3)$. The base velocities in SWA are computed using the weighted pseudo-inverse approach used in \cite{englsberger2018torque}.

\subsection{Experiments}
A position-controlled walking \cite{romualdi2019benchmarking} experiment and a torque-controlled Center of Mass (CoM) sinusoidal trajectory tracking experiment \cite{nori2015icubWBC} are used for the evaluation of the estimators on the real robotic platform, in an open-loop fashion.
We first carry out the convergence analysis done in \cite{hartley2020contact} for the walking experiment, in order to compare the performance of \DLGEKF{} with the OCEKF in estimating the observable states. 
These estimators are run for $25$ trials with the same measurements, noise parameters and prior deviations but with random initial orientations and linear velocities. 

The roll and pitch Euler angles for setting the initial IMU orientation were uniformly sampled from $-30 \si{\degree}$ to $30 \si{\degree}$, while the initial IMU linear velocities were sampled uniformly from $-0.5$ to $0.5 \si{\meter\per\second}$. Table \ref{table:parameters} shows the noise parameters and initial state standard deviations used for the experiment. 

\vspace{0.5mm}
\begin{table}
	\caption{Noise parameters and prior deviations}
	\label{table:parameters}
	\begin{center}
		\tabcolsep=0.02cm
		\scalebox{0.8}{
		\begin{tabular}{cc}	
						\begin{tabular}{c|c}
						\hline 
							Sensor & noise std dev. \\
								\hline	
							Lin. Accelerometer & 0.09 \si{\metre \per {\second^2}}	 \\
							Gyroscope & 0.01 \si{\radian \per \second} \\
							Acc. bias & 0.01 \si{\metre \per {\second^2}} \\
							Gyro. bias &  0.001 \si{\radian \per \second} \\
							Contact foot lin. velocity & 0.009 \si{\meter\per\second}\\
							Contact foot ang. velocity & 0.004 \si{\radian \per \second} \\							
							Joint encoders & 0.1 \si{\degree}						\\	\hline 
						\end{tabular}
&
					\begin{tabular}{c|c}
					\hline 
						State element & initial std dev. \\
						\hline	
						IMU \& feet position & 0.01 \si{\meter} \\						
						IMU \& feet orientation & 10 \si{\degree} \\
						IMU linear velocity & 0.5 \si{\meter\per\second} \\
						Acc. bias & 0.01 \si{\metre \per {\second^2}} \\
						Gyro. bias & 0.002  \si{\radian \per \second}
						\\	\hline 
					\end{tabular}						
		\end{tabular}
		}
	\end{center}
	\vspace{-6mm}
\end{table}

The comparison of the estimates for OCEKF and \DLGEKF{} is shown in Figure \ref{fig:random_initialstates_real}. 
\DLGEKF{} is seen to converge considerably faster than the OCEKF towards the ground truth measurements. 
The difference between the OCEKF and \DLGEKF{} lies mainly in the uncertainty representation. The former uses non-interacting manifolds for all the state elements, while the latter uses interacting manifolds (such as $SE_2(3)$ and $SE(3)$) resulting in different tangent parametrizations for the error state.
This causes the uncertainty representation to be more involved for the latter as seen in \eqref{eq:perturbedmean} resulting in more appropriate Kalman gain computations for innovation updates.

\begin{figure}
	\vspace*{-1em}
	\begin{subfigure}[b]{0.49\textwidth}
		\centering

\includegraphics[width=\textwidth]{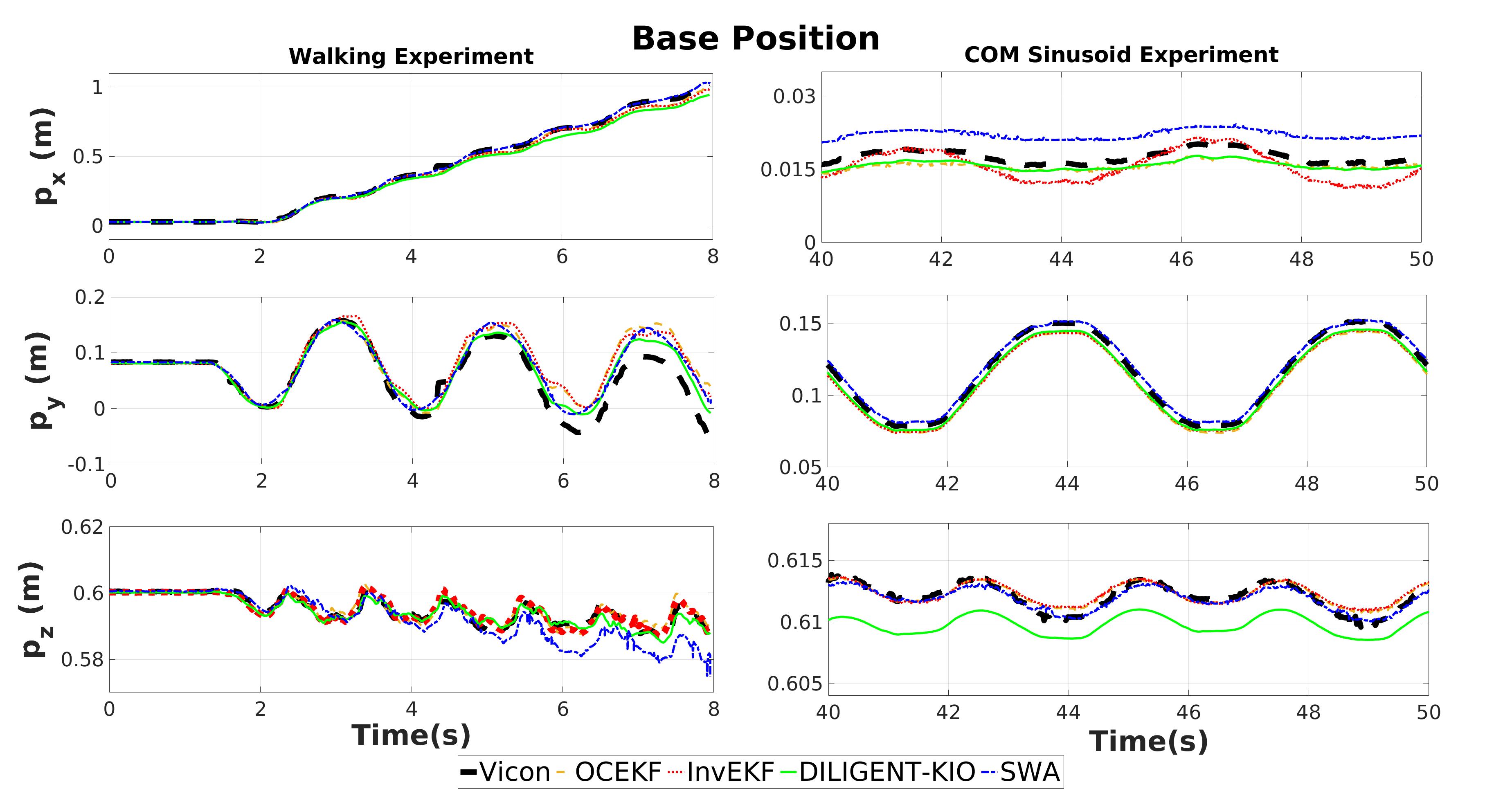}
	\end{subfigure}
\caption{\label{fig:basepos}Base position for walking (\emph{left}) and COM sinusoid experiment (\emph{right}).}

\begin{subfigure}[b]{0.49\textwidth}
	\centering
\includegraphics[width=\textwidth]{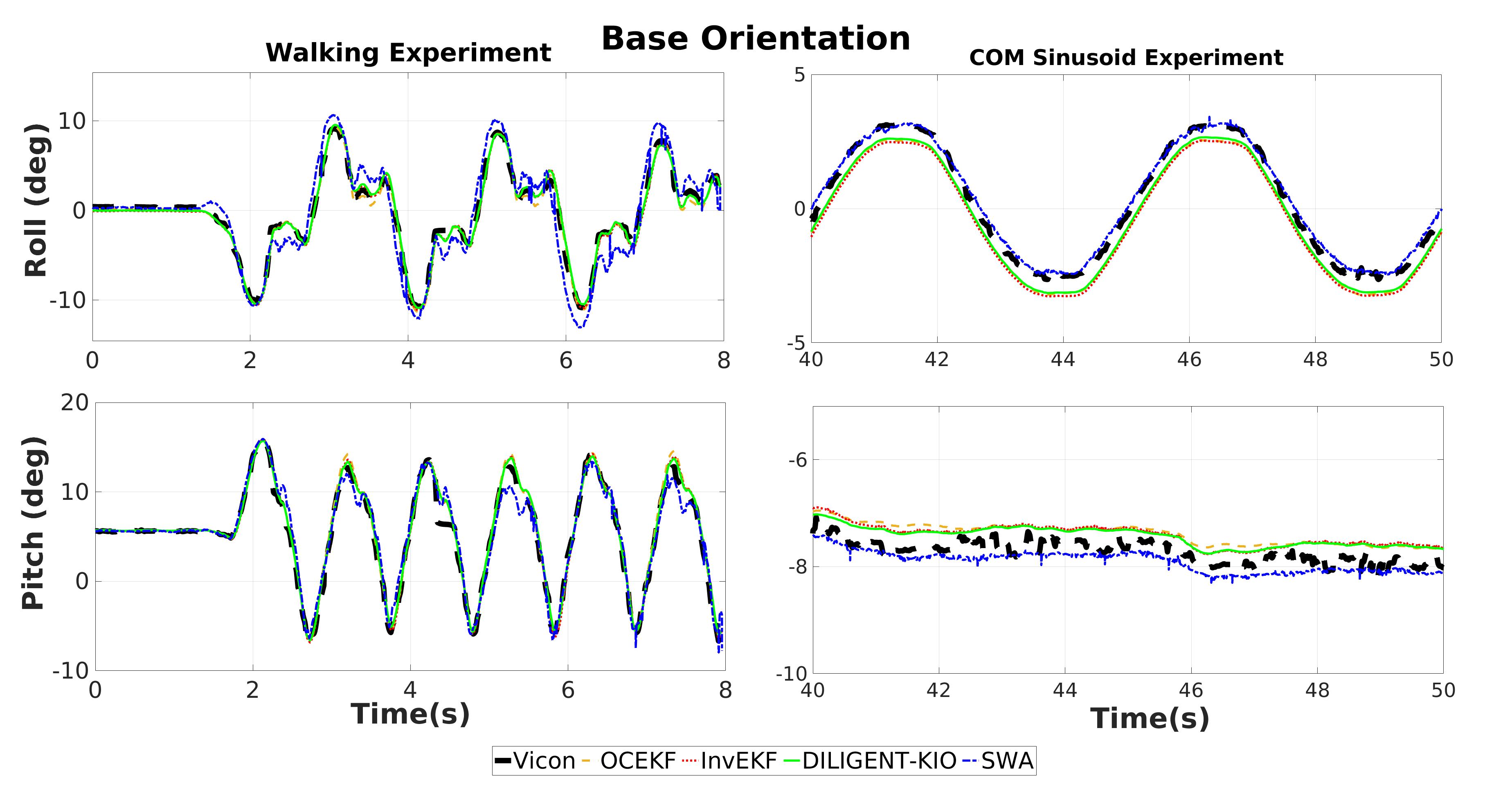}
\end{subfigure}	
    \caption{\label{fig:baserot} Base roll and pitch for walking and COM sinusoid experiment. \looseness=-1}
\vspace{-8mm}
\end{figure}

Figures \ref{fig:basepos} and \ref{fig:baserot} show the comparison of estimates from OCEKF, InvEKF,  SWA and \DLGEKF{} against the ground truth measurements for both the walking and the CoM sinusoid trajectory experiment. 
It can be noticed in Figures \ref{fig:random_initialstates_real}, \ref{fig:basepos}, and \ref{fig:baserot}, there seems to be a discontinuity in the ground truth measurements between times $t = 4.3 \si{\second}$ and $t= 4.5 \si{\second}$ for the walking experiment,  which is an outlier caused by the occlusion of Vicon markers while the robot was walking in a cluttered environment.
The Absolute Trajectory Error (ATE) \cite{zhang2018tutorial} and Relative Pose Error (RPE) \cite{sturm2012benchmark} in the left-invariant sense are shown in Table \ref{table:errors} for a $8 \si{\second}$ walking experiment and the CoM trajectory tracking experiment. 
\DLGEKF{} is seen to perform comparably with the base-line estimators for both the experiments, especially in the observable directions of orientation and velocity. 
For longer experiment durations, \DLGEKF{} suffers more from position drifts than the other filters, while the orientation and velocity errors remain comparable. \looseness=-1

\begin{table}[ht] 
\caption{Left invariant Absolute Trajectory Error (ATE) and Relative Pose Error (RPE) comparison for walking and CoM sinusoid experiment. \looseness=-1}
\centering 
\scalebox{0.8}{
\begin{tabular}{p{1cm}  p{0.5cm}  p{0.5cm}  p{0.5cm}  p{0.5cm}  p{0.5cm}  p{0.5cm}  p{0.5cm}  p{0.5cm} p{0.5cm} p{0.5cm}} 
 												Filter	& \multicolumn{5}{c}{\emph{Walking} $8\si{\second}$} & \multicolumn{5}{c}{\emph{CoM sinusoid}} \\
 												\hline
& \multicolumn{3}{c}{ATE} & \multicolumn{2}{c}{RPE}  & \multicolumn{3}{c}{ATE} & \multicolumn{2}{c}{RPE} \\
\hline
& rot [$\si{\degree}$] & pos [$\si{\meter}$] & vel [$\si{\meter/\second}$] & rot [$\si{\degree}$] & pos [$\si{\meter}$] & rot [$\si{\degree}$] & pos [$\si{\meter}$] & vel [$\si{\meter/\second}$] & rot [$\si{\degree}$] & pos [$\si{\meter}$] \\
\hline
\footnotesize{SWA} & 3.98 & \textbf{0.025} &  0.291 & 3.93 & 0.029 & \textbf{0.39} & \textbf{0.005} & 0.0545 & 0.28 & \textbf{0.0013} \\
\footnotesize{InvEKF} & 3.34 & 0.038 &  0.133 & \textbf{1.90} & \textbf{0.025} & 4.27 & 0.006 & 0.0098 & 1.36 & 0.0018  \\
\footnotesize{OCEKF} & 4.67 & 0.038 & 0.132 &  4.47 & 0.035 & 0.68 & 0.006 & 0.0094 & 0.18 & 0.0019  \\
\scriptsize{\DLGEKF{}}  & \textbf{2.29} & 0.040 &  \textbf{0.130} & \textbf{1.90} & 0.039 & 0.59 & 0.005 & \textbf{0.0089} & \textbf{0.16} & 0.0016 \\
\hline
\end{tabular} 
}
\label{table:errors} 
\vspace{-5mm}
\end{table}
\rowcolors{0}{}{}


\section{Conclusion}
\label{sec:CONCLUSION}
A proprioceptive floating base estimation was proposed using extended Kalman filtering on matrix Lie groups by considering the evolution of the state and the observations over distinct Lie groups.
The proposed filter was shown to perform better, in terms of convergence, than the   observability-constrained quaternion-based extended Kalman filter (flat foot filter).
The latter is also a discrete EKF on Lie groups, however differs from the proposed estimator in terms of uncertainty propagation.
It must be noted that the theory of autonomous error propagation \cite{barrau2016invariant, hartley2020contact} and update \cite{phogat2020discrete} or observability-based rules \cite{bloesch2013state, huang2010observability} were not \emph{explicitly} exploited within this framework. 
This could \emph{potentially} lead to inconsistencies in the filter.
Nevertheless, the proposed estimator exhibits strong convergence properties with a large basin of attraction. Future work includes fusion of the proposed estimator with exteroceptive sensor information.
\vspace{-0.3em}


\appendix
The expression for the adjoint matrix $\gadj{SE(3)}(.)$ and the closed-form expression for the left Jacobian $\gjac{SE(3)}(.)$ can be found in \cite[Appendix A]{barfoot2014associating}. 
The adjoint matrix of $SE_2(3)$   is defined as $\footnotesize \gadj{SE_2(3)}({X_\text{base}}) =\begin{bmatrix} {R} & S({p}){R} & 0_3 \\ 0_3 & {R} & 0_3 \\ 0_3 & S({v}){R} & {R}\end{bmatrix}$.
Further, we will use the definition of ${Q(\rho, \phi)}$ from \cite[Appendix A]{barfoot2014associating} to define the left Jacobian of $SE_2(3)$ as, $ \footnotesize \gjac{SE_2(3)}({\epsilon}_\text{base}) =\begin{bmatrix} {J}({\epsilon}_{R}) & {Q}({\epsilon}_{p}, {\epsilon}_{R}) & 0_3 \\ 0_3 & {J}({\epsilon}_{R}) & 0_3 \\ 0_3 & {Q}({\epsilon}_{v}, {\epsilon}_{R}) & {J}({\epsilon}_{R})\end{bmatrix}$.


\addtolength{\textheight}{-11cm}   
\bibliography{bibliography}
\bibliographystyle{IEEEtran}


\end{document}